\documentclass[11pt]{article}
\usepackage{coling2018}
\usepackage{times}
\usepackage{url}
\usepackage{latexsym}
\usepackage[export]{adjustbox}
\usepackage{xcolor}
\usepackage{amsmath,amssymb}
\usepackage{graphicx}
\usepackage[disable]{todonotes}\presetkeys{todonotes}{inline}{}
\usepackage{multirow}
\usepackage{caption}
\usepackage{subcaption}
\usepackage{booktabs}
\slname{German}

\newcommand{\tieconcat}{%
  \mathbin{\mathpalette\dotieconcat\relax}%
}
\newcommand{\dotieconcat}[2]{% auxiliary macro, don't use it directly
  \text{\raisebox{.8ex}{$\smallfrown$}}%
}
\title{Multimodal Grounding for Language Processing}
\sltitle{Multimodale konzeptuelle Verankerung f\"ur die automatische Sprachverarbeitung}
\author{Lisa Beinborn$^{\circ*}\Diamond$  \qquad Teresa Botschen$^{*}\triangle$ \qquad Iryna Gurevych $\triangle$
\vspace{5pt}\\
$\Diamond$ Language Technology Lab, University of Duisburg-Essen \vspace{3pt}\\
$\triangle$ Ubiquitous Knowledge Processing Lab (UKP) and Research Training Group AIPHES\\ 
Department of Computer Science, Technische Universit\"at Darmstadt \\
  \tt www.ukp.tu-darmstadt.de }

\begin{document}
\maketitle

\renewcommand*{\thefootnote}{\arabic{footnote}}
\begin{abstract}
This survey discusses how recent developments in multimodal processing facilitate conceptual grounding of language. We categorize the information flow in multimodal processing with respect to cognitive models of human information processing and analyze different methods for combining multimodal representations. Based on this methodological inventory, we discuss the benefit of multimodal grounding for a variety of language processing tasks and the challenges that arise. We particularly focus on multimodal grounding of verbs which play a crucial role for the compositional power of language.

\end{abstract}
\makesltitle
\begin{slabstract}
Dieser \"Uberblick er\"ortert, wie aktuelle Entwicklungen in der automatischen Verarbeitung multimodaler Inhalte die konzeptuelle Verankerung sprachlicher Inhalte erleichtern k\"onnen. Die automatischen Methoden zur Verarbeitung multimodaler Inhalte werden zun\"achst hinsichtlich der zugrundeliegenden kognitiven Modelle menschlicher Informationsverarbeitung kategorisiert. Daraus ergeben sich verschiedene Methoden um Repr\"asentationen unterschiedlicher Modalit\"aten miteinander zu kombinieren. 
Ausgehend von diesen methodischen Grundlagen wird diskutiert, wie verschiedene Forschungsprobleme in der automatischen Sprachverarbeitung von multimodaler Verankerung profitieren k\"onnen und welche Herausforderungen sich dabei ergeben. Ein besonderer Schwerpunkt wird dabei auf die multimodale konzeptuelle Verankerung von Verben gelegt, da diese eine wichtige kompositorische Funktion erf\"ullen. 
\end{slabstract}

\section{Introduction}
\label{intro}
\blfootnote{
 \hspace{-0.65cm} $^{\circ}$ The work by Lisa Beinborn has been carried out during her affiliation with Ubiquitous Knowledge Processing Lab (UKP) and Research Training Group AIPHES, Technische Universit\"at Darmstadt.
}
\blfootnote{
 \hspace{-0.65cm}    $^{*}$ The first and the second authors contributed equally to this work.}

\blfootnote{
    \hspace{-0.65cm}  % space normally used by the marker
     This work is licensed under a Creative Commons 
     Attribution 4.0 International License.
    License details:
    \url{http://creativecommons.org/licenses/by/4.0/}
}
Natural languages are continually developing constructs that include numerous variations and irregularities. 
Modeling the subtleties of language in a formal, processable way has driven computational linguistics for decades. In recent years, distributional approaches have become the most widely accepted solution to model the associative character of word meaning \cite{Harris1954,Collobert2011,Mikolov2013,Pennington2014}. These approaches learn word representations in a high-dimensional vector space based on context patterns in large text collections. Machine learning researchers aim at reducing external knowledge to an absolute minimum and simply interpret language as a continuous stream of characters. From an engineering perspective, these data-driven approaches are highly attractive because they reduce the need of domain experts. 

From a cognitive perspective, processing language in isolation without information on situational context seems to be an overly artificial setup. Human acquisition of semantic representations does not occur based on pure language input. The term conceptual grounding refers to the idea that language is grounded in perceptual experience and sensorimotor interactions with the environment \cite{Barsalou2008}. In its strictest interpretation, this embodied perspective implies that language production and language comprehension involve perceptual and motor simulations of the described situation \cite{Goldman2006}. 
An impressive number of recent neuroimaging studies indicate that processing a word activates areas in the brain that correspond to the associated sensory modality of its semantic category: action-related words like \textit{kick} trigger activity in the motor cortex and object-related words like \textit{cup} activate visual areas \cite{Pulvermueller2005,Garagnani2016}.
While it remains a controversial question to what extent conceptual representations are actually shared across modalities \cite{Louwerse2011,Leshinskaya2016}, it has been widely accepted that conceptual and sensorimotor representations are tightly coupled and interact with each other. Cognitively plausible language processing should thus interpret language as one modality within a multimodal environment.
 
This survey discusses how recent developments in multimodal processing facilitate conceptual grounding of language. It intents to provide a bridge between the field of multimodal machine learning \cite{Baltrusaitis2017} and the cognitive theories for grounding distributional semantics \cite{Baroni2016}. As this is a wide interdisciplinary topic which influences many subfields, we focus on multimodal grounding for computational linguistics. 
For a better understanding of the interaction between modalities, we categorize multimodal tasks according to the information flow between the modalities. In a second step, we analyze different methods for combining multimodal information. 
Based on this methodological inventory, we discuss the benefit of multimodal grounding for a variety of language processing tasks. In multimodal processing, grounding is usually limited to concrete concepts leading to a reduction of referential ambiguity. We provide a detailed analysis of the challenges that arise when multimodal grounding is extended to open-domain language settings. We particularly focus on multimodal grounding of verbs which is essential for the interpretation of sequences and the identification of relations between concepts.  

\section{Multimodal processing models}\label{mo}
The term ``multimodal'' has been used in a broad range of different interpretations even in the computational linguistics literature alone. In the common interpretation, modalities refer to sensory input such as audio, vision, touch, smell, and taste. Other definitions stretch over different communicative channels such as language and gesture, or simply different ``modes'' of the same modality (e.g., day and night pictures). 
In this section, we analyze the flow of multimodal information in different multimodal tasks exemplified by three modalities: natural language encoded as texts, visual signals encoded as images or videos, and audio signals encoded as sound files. For an overview of the challenges and machine learning methods associated with each task, the interested reader is referred to \newcite{Baltrusaitis2017}. We propose a classification of multimodal tasks with respect to the information flow between modalities into cross-modal transfer, cross-modal interpretation, and joint multimodal processing. 
From a historical perspective, progress in multimodal processing can be aligned with cognitive theories of multimodal organization in the human brain. 
\begin{figure}
\centering
  \begin{adjustbox}{minipage=\linewidth,scale=0.9}
  \begin{subfigure}[t]{0.3\textwidth}
  \centering
    \includegraphics[width=0.9\textwidth,height=2cm,keepaspectratio]{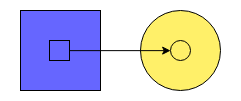}
    \caption{Cross-modal transfer. Information from modality $A$ is aligned to comparable information in $B$. }
    \label{fig:f1}
  \end{subfigure}
  \hfill
  \begin{subfigure}[t]{0.3\textwidth}
  \centering
    \includegraphics[width=0.9\textwidth,height=2cm,keepaspectratio]{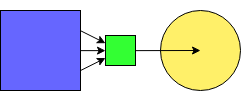}
    \caption{Cross-modal interpretation. Relevant information in modality $A$ is summarized and interpreted in modality $B$.}
    \label{fig:f2}
  \end{subfigure}
    \hfill
  \begin{subfigure}[t]{0.3\textwidth}
  \centering
    \includegraphics[width=0.9\textwidth,height=2cm,keepaspectratio]{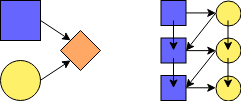}
    \caption{Joint multimodal processing. Left: Modality $A$ and $B$ both contribute to a joint prediction. Right: Interactive exchange of information between modalities.}
    \label{fig:f3}
  \end{subfigure}
      \end{adjustbox}
  \caption{Information flow in multimodal tasks. Blue and yellow shapes refer to modality $A$ and $B$.}
  \label{flow}

\end{figure}

\subsection{Cross-modal transfer}
In the 1980s and 90s, cognitive processing theories were heavily influenced by the theory of the modularity of mind \cite{Fodor1985}. It assumes that processing occurs in domain-specific encapsulated modules that do not interact with each other. Earlier approaches to multimodal engineering have taken a similar modular perspective. They model the information flow in each modality separately and the final outcome is then transferred or aligned to another modality. We group tasks in which one modality serves as the interface to query or represent the content from another modality under the category of cross-modal transfer, see Figure~\ref{fig:f1}. 

A classical example for cross-modal transfer are search and retrieval tasks. The human user provides a natural language description to query an artifact (i.e., an image, video, or audio file) from a database \cite{Atrey2010}. The cross-modal alignment between the query and the artifact requires query expansion and disambiguation for referential indexing. In speech-related transfer tasks, textual content needs to be mapped to audio samples. Speech synthesis transforms text into artificially generated phonemes for users who cannot read \cite{Zen2009}. The reverse task of transcribing audio and video content is addressed by approaches for speech recognition \cite{Juang2005} and subtitle generation \cite{Daelemans2004}. For lipreading tasks, mute video input of people speaking is transformed into text representing their utterances \cite{Ngiam2011}. 

In these cross-modal transfer tasks, synchronous processing of the input in one modality is not directly influenced by information from the output modality. The main challenge lies in finding appropriate translations or alignments from one modality to the other. Information from the output modality is mainly used for reranking of input hypotheses. This view corresponds to mental models of a language hub in the brain that does not directly incorporate perceptual information \cite{Chomsky1986}.

\subsection{Cross-modal interpretation}\label{sec:interpretation}
In order to explain how humans can select relevant information from perceptual input, the concept of attention has become very popular. \newcite{Bridewell2016} argue that attention serves as "a bottleneck for information flow in a cognitive system" that redirects mental resources. In multimodal processing, the concept of attention as a mediator between modalities is relevant for cross-modal interpretation. For these tasks, the goal is to obtain a compressed and structured intermediate representation of the input to generate a useful interpretation in the target modality.  Attention mechanisms \cite{Bahdanau2014} are used for the identification of relevant information, see Figure~\ref{fig:f2}. 

A textual interpretation of a visually presented scene is generated in image captioning \cite{Xu2015} and sketch recognition \cite{Li2015}. The goal is to identify relevant elements, group individual elements to semantic concepts, identify relations between concepts, and express these relations in natural language. The output sequence is generated while paying attention to different salient areas in the image. To our knowledge, a bidirectional information flow that includes cues from the language generation module in the image recognition process has not yet been implemented.
However, semantic information could help to better direct the attention for image recognition, e.g., the generation of a verb like \textit{eat} could constrain the visual recognition to edible objects as filler roles. \newcite{Yatskar2016} propose the task of situation recognition to approximate this problem.

Complementary approaches attempt to generate visual representations to summarize documents and present the most relevant information in an intuitive way \cite{Kucher2015}. The most popular approach are so-called word clouds which are frequency-based visualizations for topic modeling \cite{Bateman2008}. More recent approaches include semantic relations between words for a more conceptual-driven interpretation \cite{Xu2016}. Concept maps highlight structural relations between concepts in a graph-based visualization \cite{Zubrinic2012}. 
One key challenge for cross-modal interpretation tasks lies in the evaluation of the output because interpretations are by definition subjective and divergent solutions can be equally valid. Accumulations over various human ratings are currently considered to be better quality approximations than any automatic metrics \cite{Vedantam2015}.

\subsection{Joint multimodal processing}
Due to a wave of experimental findings that support the cognitive theory of embodied processing, the separating aspects between different modalities have become blurred \cite{Pulvermueller2005}. A similar development can be observed in multimodal machine learning. Tasks which explicitly require the combination of knowledge from different modalities gave rise to joint multimodal processing (Figure~\ref{fig:f3}). For emotion recognition \cite{Morency2011} or persuasiveness prediction \cite{Santos2016}, the actual content of an utterance and paraverbal cues (e.g., pitch, facial expression) need to be jointly evaluated. An ironic tone of voice might reverse the conceptual interpretation of the language content.
  
Recent work from the vision community goes one step further and tackles tasks that imperatively require an interactive flow of information. In visual question answering, a human user can ask questions about an image that the system should answer \cite{Malinowski2015}. This requires several steps: understanding the question, determining the salient elements in the image, interpreting the image with respect to the question, and generating a coherent natural language answer that matches the question. For this task, exchange of information between the modalities is crucial. In an overview, \newcite{Wu2017} compare 29 approaches to visual question answering. 23 of these approaches use a joint representation of textual and visual information. The remaining 6 approaches organize the exchange either through a coordinated network architecture or through shared external knowledge bases. 
Novel interactive approaches make it possible to directly modulate the information flow in one modality by input from another modality \cite{Devries2017} or by human feedback \cite{Ling2017}. 

The main challenge for joint processing lies in efficiently combining information from the modalities, so that redundant information is integrated without losing complementary cues. In human language understanding, this process seems to be performed in an effortless and highly accurate manner \cite{Crocker2010}. However, the underlying mechanisms of multimodal representations remain poorly understood. In Section~\ref{rep}, we discuss different methods for obtaining joint representations computationally.

\section{Multimodal representation learning}\label{rep}
Multimodal representations combine information from separate modalities. We discuss methods for representing known concepts, projecting information to represent unknown concepts, and for combining concept representations into compositional representations for sequences. 

\begin{figure}
\centering
\begin{adjustbox}{minipage=\linewidth,scale=0.9}
  \begin{subfigure}[t]{0.3\textwidth}
  \centering
    \includegraphics[width=0.9\textwidth,height=2.5cm,keepaspectratio]{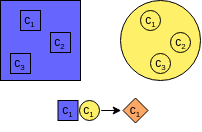}
    \caption{Multimodal fusion. Concatenate known representations from modality $A$ and $B$ and apply dimensionality reduction.}
    \label{fig:r1}
  \end{subfigure}
  \hfill
  \begin{subfigure}[t]{0.3\textwidth}
  \centering
    \includegraphics[width=0.9\textwidth,height=2.5cm,keepaspectratio]{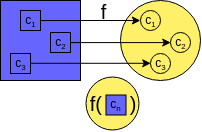}
    \caption{Mapping. Learn a mapping function $f$ from modality $A$ to $B$ that can be applied on unknown concepts $c_n$.}
    \label{fig:r2}
  \end{subfigure}
    \hfill
  \begin{subfigure}[t]{0.3\textwidth}
    \centering
    \includegraphics[width=0.9\textwidth,height=2.5cm,keepaspectratio]{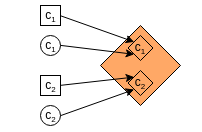}
    \caption{Joint learning. Optimize two objectives simultaneously: quality of unimodal representations and cross-modal alignment.}
    \label{fig:r3}
  \end{subfigure}
    \end{adjustbox}
  \caption{Methods for learning multimodal representations. Blue and yellow shapes indicate the representation space of modality A and B.}
  \label{reps}

\end{figure}
\subsection{Concept representations}
Even in unimodal tasks, researchers experiment with many different variations of representing concepts and their relations.
Earlier work on multimodal representations used human-elicited visual features \cite{Silberer2012,Roller2013}. 
Conveniently, integrating knowledge from different modalities has been facilitated due to the now common low-level representations of the input (also known as embeddings). Images are represented as groups of pixels, videos as series of image frames, audio data as windows of waveform samples, and language as sequences of distributional word representations. These values are then fed into a neural network that learns to compress and normalize the representation such that it better generalizes across input samples \cite{KielaBottou2014}. 
A concept representation is usually obtained by averaging over many different samples for the concept (e.g., the concept \textit{bird} is represented by averaging over the representation for $n$ images showing a bird). The representations are expressed as high-dimensional matrices which can be projected into a common space. This approach facilitates a joint information flow between different modalities and has contributed to the growing success of multimodal processing.

\paragraph{Fusion}
The most intuitive approach is multimodal fusion (Figure \ref{fig:r1}). Assuming that a unimodal vector representation $v$ for the concept $c$ and the modalities $m_1$ and $m_2$ exists, the multimodal representation $v_{mm}$ consists of the concatenation $\tieconcat$ of the two vectors weighted by a tunable parameter $\alpha$: \\
\indent \indent $v_{mm}(c) = \alpha \cdot v_{m_1}(c) \; \tieconcat \; (1 -\alpha) \cdot v_{m_2}(c) $.\\
\noindent The concatenation occurs directly on the concept level and is thus called feature-level fusion or early fusion \cite{Leong2011,Bruni2011}. In the case of pure concatenation, the unimodal representations reside in separate conceptual spaces. The concatenated representation for \textit{cat} could give us the information that \textit{cat} is visually similar to \textit{panther} and textually similar to \textit{dog}, but it is not possible to determine cross-modal similarity. In order to smooth the concatenated representations while maintaining multimodal correlations, dimensionality reduction techniques such as singular value decomposition \cite{Bruni2014} or canonical correlation analysis \cite{Silberer2012} have been applied.

\subsection{Projection}\label{zero}
In practice, concepts that have a representation in one modality are not necessarily covered by representations in another modality. The projection of unseen concepts is known as zero shot learning. It can either be performed on a mapped or a joint representational space. 
  
\paragraph{Mapping}
To overcome the lack of representations for one modality, several researchers proposed to map representations from one modality to the other (Figure~\ref{fig:r2}). The idea is to learn a mapping function $f$ from $m_1$ to $m_2$ that maximizes the similarity between a known representation of $c$ in $m_2$ and its projection from the representation in $m_1$: $c_{m_2} \sim f(c_{m_1}) $. \\
The choice of the similarity and the loss measures for learning the mapping function vary. A max-margin optimization which maximizes the similarity between true pairs of concept representations $(c_{m_1},c_{m_2})$ and minimizes the similarity for pairs with random target representations $(c_{m_1},random_{m2})$ has been shown to be a good choice for image labeling \cite{Frome2013}. In this task, the mapping approach is applied in the image-to-text direction to classify unknown objects in images based on their semantic similarity to known objects \cite{Socher2013zero} . \newcite{Lazaridou2014} and \newcite{Collell2017} proceed in the reverse text-to-image direction to ground words in the visual world. Similar propagation approaches had already been examined by \newcite{Johns2012} and \newcite{Hill2014}, but they used human-elicited perceptual features from the McRae dataset \cite{McRae2005} instead of automatically derived image representations.

\paragraph{Joint learning}
The mapping approaches assume a directed transformation from one modality to the other. Joint estimation approaches aim to learn shared representations instead (Figure~\ref{fig:r2}).  An approach inspired by topic modeling interprets aligned data as a multimodal document and uses Latent Dirichlet Allocation to derive multimodal topics \cite{Andrews2009,Feng2010,Silberer2012,Roller2013}. Unfortunately, this approach cannot be easily used for zero shot learning. 
\newcite{Lazaridou2015} enrich the skip-gram model by \newcite{Mikolov2013} with visual features. Their model optimizes two constraints: the representation of concepts $c$ with respect to their textual contexts (unsupervised skip-gram objective in $m_1$) and the similarity between word representations and their visual counterparts (supervised max-margin objective for $(c_{m_1}, c_{m_2})$). In their approach, the visual representations remain fixed, but the textual representations are learned from scratch. \newcite{Silberer2014} go one step further and use stacked multimodal autoencoders to simultaneously learn good representations for each modality (unsupervised reconstruction objective for $m_1$ and $m_2$) and their optimal multimodal combination (supervised classification objective for $(c_{m_1}, c_{m_2})$). Both approaches implicitly also learn a mapping between the two modalities and can be adjusted to induce a directional projection for zero shot learning. 
Joint learning of multimodal representations is very popular in the vision community \cite{Karpathy2014,Srivastava2012,Ngiam2011}. 

\subsection{Compositional representations}\label{compositionalRep}
For tasks that require representing longer sequences, a na\"ive approach is sequence-level fusion. In this setting, the unimodal sequence representation is obtained by performing an arithmetic operation (e.g., average, max) over the concept representations for each word in the sequence. Multimodal fusion is then performed on this averaged representation \cite{Glavas2017,Bruni2014}. \newcite{Shutova2016} work with short phrases consisting of two words and directly learn phrase representations. Missing concept representations in one modality can be obtained by mapping functions \cite{Botschen2018}. 

For image captioning approaches, representations for a pair of an image and the corresponding caption are learned jointly \cite{Kiros2014,Socher2014grounded}. Pre-trained unimodal representations are fed into a neural network which is trained with  the max-margin objective to distinguish between true and false captions for an image. The multimodal sequence representation can be obtained from the last hidden layer of the network. The introduction of attention variables can function as a mediator between the visual and the textual modality (see Section \ref{sec:interpretation}).  For a more detailed overview of multimodal sequence representations in the vision community, the interested reader is referred to \newcite{Wu2017}. The approaches for compositional representations have focused on enriching noun and adjective meaning multimodally. The multimodal interpretation of verbs as an integral part of compositional sequences has not yet been thoroughly examined.

\section{Multimodal grounding for language processing}
\label{languageProcessing}
The progress in joint multimodal processing and the increasing availability of multimodal datasets and representations open up new possibilities for grounded approaches to language processing. We review recent works for grounding concepts, grounding phrases, and grounding interaction. The challenges that arise from these efforts are discussed in Section~\ref{challenges}.

\subsection{Grounding concepts}
Multimodal concept representations are motivated by the idea that semantic relations between words are grounded in perception. Being able to assess semantic relations between concepts is an important prerequisite for modeling generalization capabilities in language processing. The combination of the textual and the visual modality has received most attention for conceptual grounding, but perceptual information from the auditory and the olfactory channel have also been used for dedicated tasks \cite{Kiela2015,Kiela2017}. In order to provide a more concrete discussion, we focus on the combination of textual and visual cues for the remainder of the survey.
 
The quality of concept representations is commonly evaluated by their ability to model semantic properties. Different approaches to learning conceptual models are compared by their performance on similarity datasets, e.g., \textit{WordSim353} \cite{Finkelstein2002}, \textit{SimLex-999} \cite{Hill2015}, \textit{MEN} \cite{Bruni2012}, \textit{SemSim}, and \textit{VisSim} \cite{Silberer2014}). These datasets contain pairs of words that have been annotated with similarity scores for the two concepts. 
Several evaluations of semantic models have
shown that multimodal concept representations outperform unimodal ones 
\cite{Feng2010,Silberer2012,Bruni2014,Kiela2014}. \newcite{Kiela2016} perform a comparison of different image sources and architectures and their ability to model semantic similarity. 
Despite the advantages of multimodal models in capturing semantic relations, it remains an open question whether they contribute to a cognitively more plausible approximation of human conceptual grounding. \newcite{Bulat2017} and \newcite{Anderson2017} conduct experiments to label brain activity scans by human subjects with the corresponding concepts that elicited the brain activity. They compare different distributional semantic models and obtain mixed results. \newcite{Bulat2017} find that visual information is beneficial for modeling concrete words, whereas \newcite{Anderson2017} conclude that textual models sufficiently integrate visual properties. Further interdisciplinary research involving computer science, neuroscience, and psycholinguistics is required to obtain a deeper understanding of cognitively plausible language processing \cite{Embick2015}. 

\subsection{Grounding phrases} 
Most experiments for conceptual grounding indicate that providing a multimodal representation for abstract concepts is significantly more challenging due to the lack of perceptual patterns associated with abstract words \cite{Hill2014tacl}. For grounding phrases, the meaning for concrete and abstract concepts need to be combined (see Section~\ref{compositionalRep}). \newcite{Bruni2012} examine the compositional meaning of color adjectives and find that multimodal representations are superior in modeling color. However, they fail to distinguish between literal and non-literal usage of color adjectives (e.g., \textit{green cup} vs \textit{green future}). 

Vivid imagery and synaesthetic associations play an important role in the interpretation of figurative language. In their influential theory of metaphor, \newcite{Lakoff1980} argue that abstract concepts can be grounded metaphorically in embodied and situated knowledge. They assume that metaphors can be understood as a mapping from a concrete source domain to a more abstract target domain. For example, \textit{time} is often viewed as a \textit{stream} that \textit{flows} in a direction. \newcite{Turney2011} operationalize this theoretical account by identifying metaphoric phrases based on the discrepancy in concreteness of source and target term. 
\newcite{Shutova2016} and \newcite{Bulat2017metaphor} build on their approach and use multimodal models for identifying metaphoric word usage in adjective-noun combinations. They show that words used in a metaphorical combination (\textit{dry wit}) exhibit less similarity than words in non-metaphorical phrases (\textit{dry skin}).  
We strongly believe that progress in multimodal compositional grounding will pave the way for a more holistic understanding of figurative language processing. As a prerequisite, multimodal grounding needs to be examined beyond the representation of concrete objects (see Section~\ref{abstractWords}). Representing verbs, compositional phrases, and even full sentences by means of multimodal information has not yet been sufficiently examined.

\subsection{Grounding interaction}
The origins of grounding theories were initiated to account for situational language use and interaction. We distinguish two main scenarios for interactive language use: language learning and situational grounding of action descriptions.

\paragraph{Grounded language learning}
Language learning is deeply rooted in social interaction and initially emerges with respect to a concrete referential context \cite{Tomasello2010}. Children acquire language in interaction with their parents and foreign language learning proceeds much faster in an environment that forces the learner to interact in the foreign language \cite{Nation1990}.
Usage-based approaches to language learning that account for the frequency and the quality of the language stimulus have a long tradition \cite{Dale1948}. \newcite{Brysbaert2009} have shown that frequency information grounded in auditory and visual communicative cues can better model human processing effects than frequency information extracted from purely textual corpora. \newcite{Lazaridou2016childes} show that a multimodal distributional approach better approximates word learning from interactive child-directed input than unimodal approaches. The same model can also convincingly simulate word meaning induction by adults \cite{Lazaridou2017chimeras}. 
Psycholinguistic research indicates that conceptual mapping modulated by visual properties is not only relevant for first language acquisition, but is also used as a means to establish cross-lingual links in foreign language learning \cite{Beinborn2014readability}.  
\newcite{Bergsma2011} and \newcite{Vulic2016} take advantage of this observation and use multimodal representations to induce multilingual representations.

\paragraph{Grounding sequences in actions} 
Situational grounding of action descriptions requires the representation of sequences and their compositional interpretation. \newcite{Regneri2013} build a corpus that grounds descriptions of actions in videos showing these actions. For the interpretation of sequences, evaluating verbs and their arguments plays a fundamental role. \newcite{Yatskar2016} developed the \textit{imSitu} dataset which consists of images depicting verbs and annotations which link the verb arguments to visual referents. This dataset can be used for the multimodal task of situation recognition \cite{Mallya2017,Zellers2017}, and it serves as a multimodal resource for verb processing. Grounding verbs is particularly challenging because of the variety of their possible visual instantiations. For example, an image of an adult drinking beer has very little in common with a zebra drinking water. 
 
Multimodal interpretation of sequences is highly relevant for robotics research \cite{Singh2017}.  \newcite{Mordatch2017} examine the emergence of compositionality in grounded multi-agent communication. The language learned by artificial agents is not necessarily interpretable by humans. \newcite{Lazaridou2017agent} show that agents which develop their own language for representing concepts that are grounded in images infer similar taxonomic relations as humans. Their work suggests that the learned concepts can even be mapped back into natural language. Agent-agent communication has already been examined in the talking head experiments, in which two agents learn to discriminate between objects and develop their own language of referring expressions \cite{steels1997grounding,steels2002grounding}. \newcite{Hermann2017} and \newcite{Heinrich2018} explicitly focus on human-robot interaction and train their agent to associate natural language descriptions of actions with perceptual input from its sensors. 

For experiments on grounded language understanding, the situational environment is usually artificially restricted to a very small domain. This confined setting facilitates the analysis of compositional expressions and their referential interpretation as complex object descriptions or action sequences. 
In open-domain language understanding, semantic disambiguation is even more challenging. Approaches using multimodal information for the disambiguation of concepts \cite{Xie2017}, named entities \cite{Moon2018}, and sentences \cite{Botschen2018,Shutova2017semantic} show promising tendencies, but the underlying compositional principles are not yet understood. 

\section{Challenges for grounded language processing}\label{challenges}
Multimodal grounding of language has been a longstanding goal of language researchers. The discussion has gained new momentum due to the recent developments in learning distributed multimodal representations. Most evaluations indicate that multimodal representations are beneficial for a variety of tasks, but explanatory analyses of this effect are still in a developing phase. In this section, we discuss open challenges that arise from existing work. For future work, we propose to examine multimodal grounding beyond concrete nouns and adjectives. In order to do this, larger multimodal datasets encompassing a wider range of word classes need to be build. These datasets would enable us to analyze compositional representations in more detail and to develop more elaborate models of selective multimodal grounding. 
\subsection{Combining complementary information}
Different modalities contribute qualitatively different conceptual information. \newcite{Bruni2014} argue that highly relevant visual properties are often not represented by linguistic models because they are too obvious to be explicitly mentioned in text (e.g., birds have wings, violins are brown). Textual models, on the other hand, provide a better intuition of taxonomic and functional relations between concepts which cannot easily be derived from images \cite{Collell2016differences}. Ideally, multimodal representations should integrate the complementary perspectives for a more coherent grounded interpretation of language. From a more skeptical perspective, \newcite{Louwerse2011} states that perceptual information is already sufficiently encoded in textual cues. In this case, the superior performance of multimodal representations that has been established by several researchers would mainly be due to a more robust representation of highly redundant information. 
The results by \newcite{Silberer2014} and \newcite{Hill2014tacl} support the intuitive assumption that textual representations better model textual similarity and visual representations better model visual similarity. As the multimodal models improve on both similarity tasks, the integration of complementary information seems to be successful. Interestingly, both evaluations show that simply concatenating the two modalities already yields a quite competitive model. The reported findings have been evaluated on models working with human-annotated perceptual features. These features inherently represent taxonomic knowledge that cannot be directly inferred from visual input. It remains an open question to which extent automatically derived image representations can contribute complementary information when combined with textual representations. Most multimodal research to date focuses on the representation of individual concepts (nouns) and their properties (adjectives). The benefit of multimodal representations for language tasks going beyond concept similarity needs to be examined in more detail from both, engineering and theoretical perspectives.  

\begin{figure}
\centering
\includegraphics[scale=0.3]{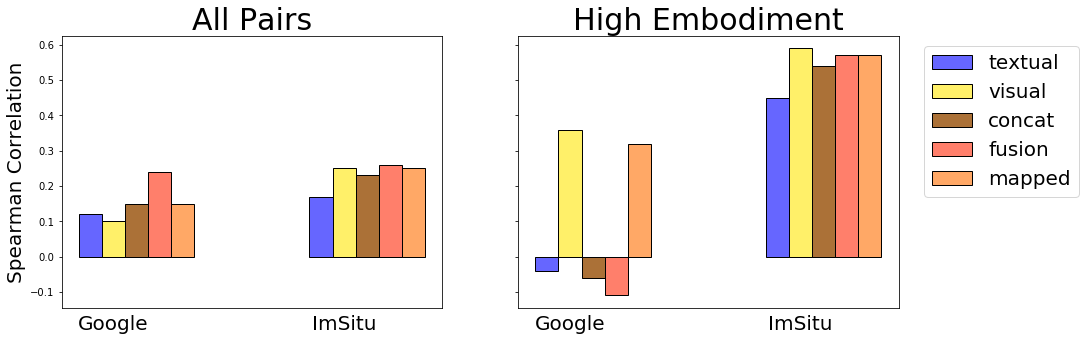}
\caption{Illustration for the quality of verb representations indicated as Spearman correlation between the cosine similarity of verbs and their corresponding similarity rating in the \textit{SimVerb} dataset.} 
\label{barplot}
\end{figure}
\paragraph{Multimodal grounding of verbs}
Verbs play a fundamental role for expressing relations between concepts and their situational functionality \cite{Hartshorne2014}. The dynamic nature of verbs poses a challenge for multimodal grounding. To our best knowledge, only \newcite{Hill2014tacl} and \newcite{Collell2017} consider verbs in their evaluation. They report that results for verbs are significantly worse, but do not elaborate on this finding. 
We present first steps towards an investigation of verb grounding.\footnote{The pre-trained embeddings and the script to reproduce our results are available for research purposes: {https://github.com/UKPLab/coling18-multimodalSurvey.}} Figure~\ref{barplot} illustrate the quality of verb representations in the most common publicly available approaches for multimodal representations. In line with previous work, the quality of the representations is evaluated as the Spearman correlation between the cosine similarity of two verbs and their corresponding similarity rating in the \textit{SimVerb} dataset \cite{Gerz2016}. We compare the quality of 3498 verb pairs\footnote{Two pairs had to be excluded because \textit{misspend} was not covered in the textual representations.} in textual Glove representations \cite{Pennington2014} and two visual datasets: the Google dataset that performed best in \newcite{Kiela2016} and has the highest coverage for the verb pairs (493 pairs, 14\%)\footnote{The coverage in \textit{WN9-IMG} \cite{Xie2017} and the dataset used by \newcite{Collell2017} is lower.} and the \textit{imSitu} dataset which has been intentionally designed for verb identification (354 pairs, 10\%). 
The results show that models which include visual information outperform purely textual representations for known concepts. However, the general quality of the verb representations is much lower than the quality reported for nouns. As a consequence, the mapping to unseen verb pairs yields unsatisfactory results for the full \textit{SimVerb} dataset. Our encouraging results for the \textit{imSitu} dataset indicate that it is recommended to directly obtain visual representations for verbs instead of projecting the meaning. Building larger multimodal datasets with a focus on verbs seems to be a promising strand of research for future work. 

\subsection{Imageability of abstract words}\label{abstractWords}
Conceptual grounding of language can be intuitively performed for concrete words that have direct referents in sensory experience. \newcite{Bruni2014} and \newcite{Hill2014} show that multimodal representations are beneficial for evaluating concrete words, but have little to no impact on the evaluation of abstract words. 
Projecting unseen concepts into the representation space based on their relations to seen concepts in another modality provides an elegant method for zero shot learning, but it is questionable whether multimodal relations between concrete concepts are sufficient to infer relations between abstract concepts. 
\newcite{Lazaridou2015} analyze projected abstract words by extracting the nearest visual neighbor from their multimodal representation. The neighbors were paired with random images and human raters judged how well each image represents the word. The hypothesis that concrete objects are more likely to be captured adequately by multimodal representations was confirmed. However, they also provide illustrating examples which represent abstract words like \textit{together} or \textit{theory} surprisingly well. 

\paragraph{Embodiment of verbs}
From a multimodal perspective, verbs can be categorized according to their degree of embodiment. This measure indicates to which extent verb meanings involve bodily experience \cite{Sidhu2014}. We obtain embodiment ratings for 1163 pairs.\footnote{\url{https://psychology.ucalgary.ca/languageprocessing/node/22}. We only include a pair, if an embodiment rating is available for both verbs.} The class \textit{high embodiment} contains pairs like \textit{fall-dive} in which the embodiment of both verbs can be found in the highest quartile (135 pairs),  \textit{low embodiment} contains pairs with embodiment ratings in the lowest quartile (81 pairs) like \textit{know-decide}.\footnote{It should be noted that not all instances of the two classes are covered by the visual representations. The small number of instances might have an impact on the correlation values.}   
Coherent with previous work on concrete and abstract nouns \cite{Hill2014tacl}, it can be seen that visual representations better capture the similarity of verbs with a high level of embodiment. The mapped representations maintain this sense of embodiment, whereas the concatenated and fused representations better capture the similarity for verbs referring to more conceptual actions. This finding indicates that multimodal information is not equally beneficial for all words. 
\subsection{Selective multimodal grounding}
The expressive power of language is essentially due to its combinatorial capabilities. Understanding how to combine concept representations to represent multi-word expressions or even full sentences has been a question of ongoing research in computational linguistics for decades. The inclusion of additional modalities further complicates this debate. 
\newcite{Glavas2017} and \newcite{Botschen2018} obtain multimodal sentence representations by averaging over the multimodal representations for each word. They report improved results for the tasks of sentence similarity and frame identification. Our comparison above indicates that this superior performance is mainly due to a better representation of concepts. This raises the assumption that multimodal grounding should only be performed on selected words. 
\newcite{Glavas2017} propose to condition the inclusion of visual information on the prototypicality of a concept as measured by the image dispersion score  \cite{Kiela2014}. This measure calculates the average pairwise cosine distance in a set of images to model the assumption that an image collection for an abstract concept like \textit{happiness} is more diverse than for a concrete concept like \textit{ladder}. \newcite{Lazaridou2015} and \newcite{Hessel2018} propose alternative concreteness measures based on the same idea. 
Unfortunately, these measures are highly dependent on the image retrieval algorithm which might be optimized towards obtaining a diverse range of images.
Nevertheless, we assume that selective multimodal grounding constitutes a more plausible approach to sentence processing. Some functional words (e.g., locative expressions) might benefit from multimodal information, but it currently remains unclear how words with syntactic functions (e.g., coordinating expressions) should be represented visually. 
\section{Conclusion}
We analyzed how multimodal processing has developed from transfer between encapsulated modalities to interactive processing over joint multimodal representations. These developments contribute to new avenues of research for grounded language processing. 
We strongly believe that the integration of multimodal information will improve our understanding of conceptual semantic models, figurative language processing, language learning, and situated interaction. Image datasets are often optimized towards providing a variety of visual instantiations. Developing algorithms for determining more prototypical visual representations could contribute to better grounding of verbs and might also serve as a criterion for selective multimodal grounding.

\section*{Acknowledgements}
This work has been supported by the DFG-funded research training group ``Adaptive Preparation of Information form Heterogeneous Sources'' (AIPHES, GRK 1994/1) at Technische Universit\"at Darmstadt.
We thank Faraz Saeedan for his assistance with the computation of the visual embeddings for the imSitu images.
We thank the anonymous reviewers for their insightful comments.
\bibliographystyle{acl}
\bibliography{ConceptualGrounding}

\end{document}